\documentclass[10pt, a4paper]{article}

\usepackage{lrec-coling2024}

\usepackage{tempora}
\usepackage[T2A,T1]{fontenc}
\usepackage[english, russian]{babel}
\addto\captionsrussian{
\renewcommand{\abstractname}{Abstract}

}

\usepackage{xcolor}
\usepackage{hyperref}
\definecolor{darkblue}{rgb}{0, 0, 0.5}
\hypersetup{colorlinks=true, citecolor=darkblue, linkcolor=darkblue, urlcolor=darkblue}
\usepackage{xstring}
\usepackage{color}

\usepackage{url}
\usepackage{tipa}

\usepackage{multirow,booktabs,url}
\usepackage{dialogue}

\usepackage{amssymb}
\usepackage{pifont}
\newcommand{\cmark}{\ding{51}}
\newcommand{\xmark}{\ding{55}}
\newcommand*{\email}[1]{
\normalsize\href{mailto:#1}{#1}\par
}

\title{KazQAD: Kazakh Open-Domain Question Answering Dataset}

\name{\begin{tabular}{c}
Rustem Yeshpanov\textsuperscript{1},
Pavel Efimov\textsuperscript{2},
Leonid Boytsov\textsuperscript{3$\star$}\\
Ardak Shalkarbayuli\textsuperscript{4},
Pavel Braslavski\textsuperscript{5} \end{tabular}}

\address{
\textsuperscript{1}Institute of Smart Systems and Artificial Intelligence, Nazarbayev University, Astana, Kazakhstan\\ 
\textsuperscript{2}ITMO University, Saint Petersburg, Russia\\
\textsuperscript{3}Amazon AWS AI Labs, Pittsburgh, USA\\
\textsuperscript{4}Suleyman Demirel University, Almaty, Kazakhstan\\
\textsuperscript{5}School of Engineering and Digital Sciences, Nazarbayev University, Astana, Kazakhstan\\
\email{rustem.yeshpanov@nu.edu.kz}, \email{pavel.vl.efimov@gmail.com}, \email{leo@boytsov.info} \\
\email{ardak.shalkar@gmail.com}, \email{pbras@yandex.ru}}

\abstract{
We introduce KazQAD---a Kazakh open-domain question answering (ODQA) dataset---that can be used in both reading comprehension and full ODQA settings, as well as for information retrieval experiments.
KazQAD contains just under 6,000 unique questions with extracted short answers and nearly 12,000 passage-level relevance judgements.
We use a combination of machine translation, Wikipedia search, and in-house manual annotation to ensure annotation efficiency and data quality.
The questions come from two sources: translated items from the Natural Questions (NQ) dataset (only for training) and 
the original Kazakh Unified National Testing (UNT) exam (for development and testing).
The accompanying text corpus contains more than 800,000 passages from the Kazakh Wikipedia.  
As a supplementary dataset, we release around 61,000 question-passage-answer triples from the NQ dataset that have been machine-translated into Kazakh.
We develop baseline retrievers and readers that achieve reasonable scores in 
retrieval (NDCG@10 = 0.389 MRR = 0.382), reading comprehension (EM~=~38.5 F1 = 54.2),
and full ODQA (EM = 17.8 F1 = 28.7) settings. 
Nevertheless, these results are substantially lower than state-of-the-art results for English QA collections, and we think that there should still be ample room for improvement.
We also show that the current OpenAI's ChatGPTv3.5 is not able to answer KazQAD test questions in the closed-book setting with acceptable quality. 
The dataset is freely available under the Creative Commons licence (CC BY-SA) at \url{https://github.com/IS2AI/KazQAD}. 
\\ \newline 
}

\begin{document}

\maketitleabstract

\section{Introduction}

\def\thefootnote{$\star$}\footnotetext{Work done outside of the scope of employment.}
\def\thefootnote{\arabic{footnote}}

Open-domain question answering (ODQA) is the task of finding a concise and accurate answer to a natural language  question in a large collection of text documents~\cite{hirschman2001natural}.
A more restricted question-answering (QA) task that involves locating an answer within a single document is commonly referred to as \emph{reading comprehension} (RC).
A traditional ODQA system has two main components: a \emph{retriever} 
and an \emph{answer extractor} ~\cite{DBLP:journals/ftir/Prager06,DBLP:journals/aim/FerrucciBCFGKLMNPSW10}.
In recent literature, answer extractors  are called \emph{readers} \cite{chen-etal-2017-reading,zhu2021retrieving}. 

QA is a popular practical application and an active research area that serves as a testbed for information retrieval (IR) and natural language processing (NLP) techniques.
The success of evaluation initiatives, such as TREC~\cite{voorhees2005trec} and CLEF~\cite{ferro2019multilingual}, as well as datasets such as SQuAD \cite{squad} 
have demonstrated that standardised evaluation approaches and test collections are key factors for measurable progress in solving IR and NLP tasks. 
However, despite significant research investment and impressive progress in English QA~\cite{CALIJORNESOARES2020635},  
advances in other languages have been less impressive.
This is in part due to the scarcity of training and test datasets, which are more difficult to create for low-resource languages~\cite{ruder2022statemultilingualai}. 

To mitigate this shortcoming, we create a new ODQA dataset \textbf{KazQAD} (/k\ae s\textprimstress ke\textsci d/), which stands for a \textbf{Kaz}akh open-domain \textbf{Q}uestion \textbf{A}nswering \textbf{D}ataset. 
\textbf{KazQAD} can be used in both reading comprehension and full ODQA settings, as well as for information retrieval experiments.

Kazakh, as a member of the Turkic language family and specifically of its Kipchak branch, is characterised as an agglutinative language~\cite{campbell2020compendium}.
Written communication in Kazakh relies on an extended Cyrillic script.
It is estimated that there are approximately 13 million native speakers of Kazakh, over 10 million of whom reside in Kazakhstan.
The remaining three million speakers are scattered across various other countries, including China, Mongolia, Russia, and Turkey.

Kazakh is considered a language with limited resources. 
Annotated datasets specifically tailored to IR and NLP tasks in Kazakh are scarce.
A notable exception is a recent dataset focusing on named entity recognition (NER)---KazNERD~\cite{yeshpanov-etal-2022-kaznerd}.

The main idea behind KazQAD is to leverage and repurpose existing data in addition to manually labeling
a new resource.
At the same time, we refrained from adopting a fully automated approach to dataset construction such as relying solely on machine translation to prevent an extensive presence of unrealistic synthetic data such as \emph{translationese}~\cite{baroni2006new}.
For the training set, we extracted questions from the English \emph{Natural Questions} (NQ) dataset~\cite{kwiatkowski2019natural}, machine-translated them into Kazakh and aligned them with the corresponding Kazakh Wikipedia articles.
For the development and test sets, we used questions from the Kazakh Unified National Testing (UNT) exam and matched them with Wikipedia pages using Google search.
Then, in-house native Kazakh  annotators extracted answers from the retrieved Wikipedia passages.
The data processing and annotation is described in detail in Section~\ref{sec:annotation}. 
Overall, KazQAD contains just under 6,000 unique questions with extracted short answers and nearly 12,000 passage-level relevance judgements.

We develop baseline retrievers and readers that achieve reasonable scores in 
retrieval (NDCG@10 = 0.389 MRR = 0.382), reading comprehension (EM = 38.5 F1 = 54.2),
and full ODQA (EM = 17.8 F1 = 28.7) settings.
Yet, these results are mostly worse than those reported by~\citet{chen-etal-2017-reading}, who described possibly the first neural ODQA English system.
Thus, we believe that there is still much room for improvement.
In addition, we submitted KazQAD test questions to ChatGPTv3.5 and evaluated its answers.
The combination of automatic and manual evaluation shows that OpenAI's model still struggles to answer factual questions in Kazakh. 
The dataset and baseline models are freely available under the Creative Commons licence.\footnote{\url{https://github.com/IS2AI/KazQAD}}

\section{Related Work}

TREC~\cite{voorhees2005trec}, 
the Cross-Language Evaluation Forum (CLEF)~\cite{ferro2019multilingual}, the Russian information retrieval evaluation initiative (also known as ROMIP)~\cite{romip2004}, and NII Testbeds and Community for Information access Research ({NTCIR})\footnote{\url{https://research.nii.ac.jp/ntcir/}}  evaluation campaigns have featured both cross-language retrieval (involving query and document collections in different languages) and monolingual retrieval in non-English languages.
However, the datasets produced by these evaluation initiatives have been relatively small. 
With the proliferation of data-intensive neural methods in IR and QA, the demand for larger annotated collections has increased significantly. 

Since the release of the English SQuAD dataset~\cite{squad}, we have experienced a ``QA dataset explosion''~\cite{rogers2023qa} that has, inter alia, led to the emergence of many non-English datasets.
These datasets were created through various approaches, including machine translations of SQuAD, such as the Spanish~\cite{spanishSQuAD} and Turkish~\cite{unlu-menevse-etal-2022-framework} variants, 
as well as the application of the SQuAD annotation approach to Wikipedia in other languages (cf. Russian SberQuAD~\cite{sberquad}).

Subsequently, multilingual QA datasets appeared, encompassing multiple languages simultaneously. 
The approaches to their creation also varied. 
For instance, XQuAD~\cite{xquad} involved manual translation of a small portion of English SQuAD questions and contexts into 10 languages, while MLQA~\cite{mlqa} focused on translating only the questions and annotating the answers in parallel contexts for each of the seven languages independently.
TyDi QA~\cite{tydiqa} utilised annotators that independently generated questions based on short prompts from Wikipedia for 11 typologically different languages, annotated them at paragraph level, and extracted short answers.
MKQA~\cite{mkqa} involved the manual translation of original English questions from NQ~\cite{kwiatkowski2019natural} into 25 languages and their subsequent annotatiion with answers from scratch.

The Kazakh language is present in the recently released massive multilingual \textsc{Belebele} dataset~\cite{bandarkar2023belebele}. 
The dataset contains 900 multiple-choice questions to 488 passages from the \textsc{FLoRes} collection~\cite{goyal2022flores}, translated from English into 122 languages, including Kazakh. 
Although KazQAD is a monolingual dataset, it has the following advantages: 
\begin{itemize}
\item It is larger;
\vspace{-0.2cm}
\item It has a much more diverse set of potential answers compared to the multiple-choice \textsc{Belebele};
\vspace{-0.2cm}
\item It uses original Kazakh questions (for testing) and passages instead of translations;
\vspace{-0.2cm}
\item It can be used in both RC and ODQA settings, as well as a testbed for IR models.
\end{itemize}.
\vspace{-0.3cm}

The English MS MARCO dataset~\cite{bajaj2016ms} and its subsequent editions \cite{DBLP:journals/corr/abs-2003-07820} have become the \textit{de facto} standard for conducting search and ranking experiments using neural models. 
MS MARCO has also been machine-translated into 13 languages~\cite{bonifacio2021mmarco}.
~\citet{chen-etal-2017-reading} and later~\citet{karpukhin2020dense} showed that extractive QA datasets can be leveraged to create datasets for search and ranking. 
Thus, the multilingual TyDi QA dataset served as the foundation for the creation of the IR dataset Mr. TyDi~\cite{mrtydi}, which covers 11 languages. 
Later, the dataset annotations were updated and extended to include data from an additional seven languages, resulting in the creation of the MIRACL dataset~\cite{zhang2022making}.

Large language models (LLMs) have the potential to revolutionise data annotation efforts, model training, and evaluation.
For example, LLMs can be leveraged for generating training examples and subsequently fine-tuning a smaller model on them~\cite{bonifacio2022inpars}.
Furthermore, LLMs can also be directly utilised for evaluation purposes~\cite{fu2023gptscore}. 
However, when it comes to low-resourced languages, these options entail certain risks.
Limited research has been conducted on the general capabilities of LLMs in such languages, beyond their application in machine translation~\cite{jiao2023chatgpt,lai2023chatgpt}. 
As we show in our study, the question answering capabilities of ChatGPT in Kazakh are still in their infancy. 

When creating KazQAD, we integrated various approaches employed in the construction of established QA datasets.
For the training set
of our dataset, we followed a similar strategy as for MKQA, where we ``recycled'' the English NQ questions while
sourcing the passages to be annotated from the corresponding Kazakh Wikipedia (which is similar to the process of creating MLQA by~\citet{mlqa}). 
In addition, we leveraged question-answer pairs to identify relevant passages for another segment of KazQAD, taking inspiration from the DPR study~\cite{karpukhin2020dense}, but the passages retrieved were annotated manually.

\section{Dataset Creation} \label{sec:annotation}

\subsection{Data Sources}
\paragraph{Kazakh Wikipedia} 
We used a dump of the Kazakh Wikipedia dated January 20, 2023, which contains 236,480 pages.
To extract the text content of the pages, we employed a combination of  WikiExtractor~\cite{Wikiextractor2015} and a custom parser specifically designed for the templates prevalent in the Kazakh Wikipedia.
We then split the pages into passages using double newlines ($\backslash$n$\backslash$n).
We found that extremely long passages (tens of thousands of characters) were still present in the collection, so we further subdivided them using single newlines ($\backslash$n).
Finally, we removed exact passage duplicates.
As a result, our collection contains 
around 815,000 passages, with an average length of 277 characters and a median of 183 characters. 
We also aggregated page views and edits for each page for the entire year 2022.
We used these data as page quality indicators in the annotation process.

\vspace{-0.3cm}

\begin{figure*}[t]
\centering
\includegraphics[width=0.80\textwidth]
{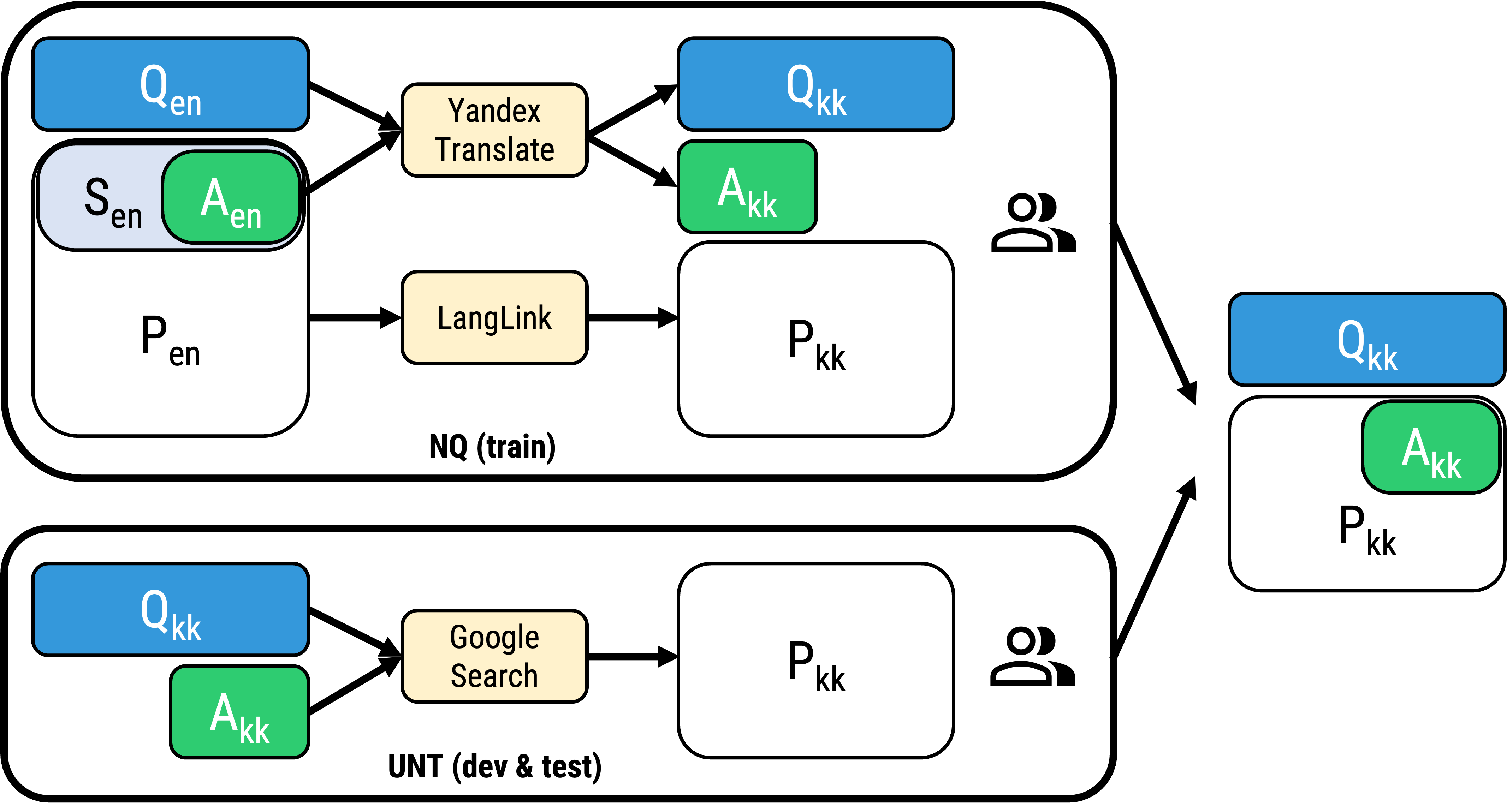}
\caption{KazQAD pipeline: \emph{Q} -- question, \emph{P} -- passage, \emph{S} -- sentence, \emph{A} -- answer; subscripts: en -- English, kk -- Kazakh. The upper part corresponds to the training set, while the lower part represents the development and test sets. In the case of the training set, we start with the pre-selected NQ questions and in-context answers machine-translated into Kazakh. Candidate passages are extracted from the parallel Kazakh Wikipedia pages. In the case of the development and test sets, the starting point is an original Kazakh question-answer pair from the UNT collection. Passages are retrieved from the Kazakh Wikipedia using Google search. In both annotation scenarios, the annotators are thus presented with question-paragraph pairs along with candidate answers.}
\label{fig:kazqad}
\end{figure*}

\paragraph{Natural Questions (NQ)} 
NQ is a dataset containing over 300,000 English questions from the Google search log, in part along with \emph{long answers} (roughly corresponding to Wikipedia paragraphs) and \emph{short answers} from the English Wikipedia~\cite{kwiatkowski2019natural}.
Short extractive answers are provided for more than 100,000 questions.
The rationale behind our choosing NQ lies in its size and notably diverse nature as a QA collection. 
Furthermore, its questions are representative of authentic user information needs, unlike datasets such as SQuAD where the questions are generated by crowdworkers presented with Wikipedia paragraphs.
For the \emph{training} split of KazQAD, we utilise machine-translated NQ questions.

\vspace{-0.3cm}

\paragraph{NQ translation} In addition, we machine-translated a substantial part of the English NQ dataset into Kazakh. 
We selected question-paragraph pairs for which there is a short answer and the passage does not belong to the English Wikipedia page for which there is a parallel Kazakh page. 
We labelled the answer in the passage and translated these data using the Yandex Translate API.\footnote{Based on our internal evaluation, Google Translate and Yandex Translate demonstrate comparable performance on the English-Kazakh pair within the FLoRes dataset, as measured by BLEU scores. The cost-effectiveness of Yandex Translate significantly influenced our decision, as it offers a more economical solution compared to Google Translate.}
In total, we ended up with more than 61,000 question-passage pairs with short answers that we used in experiments (we also make them publicly available). 

\vspace{-0.3cm}

\paragraph{Unified National Testing (UNT)} The UNT is a comprehensive school leaving (i.e., graduation) qualification  exam in Kazakhstan that covers a diverse range of subjects.
These include Kazakh language, Kazakh literature, history of Kazakhstan, mathematics, physics, biology, and several others.
Part of the test consists of multiple-choice questions.
There are many examples of UNT questions and related study materials on the Internet. 
We collected 8,582 questions along with the correct answers in five subjects (biology, geography, history of Kazakhstan, Kazakh literature, and world history). 
We instructed the annotators to rephrase the multiple-choice questions into open-ended questions where possible, which resulted in 8,562 questions.
We use UNT questions for the \emph{development} and \emph{test} splits of KazQAD.    

\subsection{Annotation}

Since we were able to collect only a limited number of original Kazakh UNT questions from the Internet, we decided to make them the basis of the development and test sets.
We chose NQ questions as the source for the training set. 
Despite the fact that these data sources have different structure, we unified the data using pre-processing, so that the annotators could solve the same problem on different subsets of the data.
For example, we machine-translated English questions and potential answers into Kazakh, and unsed Wikipedia langlinks to find candidate relevant passages. 
In the case of UNT, we found potentially relevant passages using Google search with question and answer as a query.
It is important to note that in both cases we used the same collection of original passages from the Kazakh Wikipedia.
As a result of the annotation, we obtained question-passage-answer triples (as well as passages marked as non-relevant by the annotators).
These data can be used for various tasks: searching a collection of Wikipedia passages (answer-bearing passages are considered relevant), reading comprehension (using full question-passage-answer triples), and ODQA (using question-answer pairs). 

The KazQAD annotation pipeline is outlined in Figure~\ref{fig:kazqad}.
The upper part corresponds to NQ processing resulting in the training set, while the lower part corresponds to UNT processing (the development and test sets). 
We hired six Kazakh native speakers
(three women and three men).
The age distribution of the annotators ranged from 27 to 35, with a mean age of 31. Their educational backgrounds varied, with two holding Master's degrees and the remaining four having Bachelor's degrees.

\paragraph{NQ} First, we identified NQ questions that had at least one short answer in the dataset. 
In addition, to be included, the Wikipedia page had to have a corresponding parallel Kazakh version available. 
Among the distinct Wikipedia titles in the NQ dataset, a total of 10,016 titles were identified to have corresponding Kazakh pages. 
Of the 42,451 unique questions associated with these titles, 14,553 questions were identified that contained at least one answer.
These questions, along with the answer-bearing sentences, were machine-translated into Kazakh using the Yandex translate service.\footnote{\url{https://translate.yandex.com/}} 
To be able to locate the answer in the translated sentence, we prudently placed the answer span in quotation marks in the English source, as in~\citet{mlqa}.
In about 10\% of the translated sentences, we could not locate the supposed answer. 
Several semantically similar questions from NQ were translated into identical Kazakh questions; their annotations were subsequently merged.  
Next, we extracted up to four passages from the parallel Kazakh page using Wikipedia \emph{langlinks}.
The total count of question-passage pairs amounted to 34,660, as some pages contained minimal information, often limited to a single short paragraph.
Note that we machine-translated only the questions, not the passages. 

Using the NQ data, we implemented a two-step annotation process.
In the first step, the annotators labelled passages as either relevant or non-relevant.
In the second step, they selected a short answer (or multiple answers) from the passages marked as relevant. 
If a question was found to be ambiguous or incomprehensible, the annotators had the option to discard it. 

During the annotation process, questions that pertained to events that were no longer true at the time of annotation, but could still be answered using the provided passage, were deemed ambiguous and subsequently discarded. 
For instance, if a question enquired about the current Queen of England, it was considered unclear since, at the time of annotation, the reigning monarch was King Charles III.
This decision was made even if the passage contained information regarding Queen Elizabeth II. 
Similarly, a question about the venue of the upcoming 2022 football World Cup was marked as ambiguous as the tournament had already taken place at the time of annotation, although the passage indicated that it would be held in Qatar. 
By eliminating such questions, we were able to maintain the accuracy and contemporaneity of the dataset.

\paragraph{UNT} To complement the UNT data with contexts, we utilised a combination of a UNT question and an answer as a query and employed Google custom search restricted to the \texttt{kk.wikipedia.org} domain, specifically focusing on the Kazakh Wikipedia.
We took passages from our Wikipedia collection (see above) corresponding to the top-ranked pages returned by Google search and re-ranked them using a simple heuristic of similarity to the question and answer.
The annotators were presented with up to four top-ranked passages.
It is crucial to acknowledge that, after scrutinizing the results of the NQ annotation and consulting with the annotators, we transitioned to a one-step annotation approach for the UNT data, which was anticipated to streamline the process. Under this annotation scheme, the annotators were presented with passages and answer hints obtained through machine translation. Their task was to either identify short answers or declare absence thereof. When an annotator found a short answer in a passage, they proceeded to the next question. This difference in the annotation scheme accounts for the varying ratios of relevant and non-relevant labels and question-paragraph pairs with or without answers between the NQ and UNT parts. 

\subsection{KazQAD Structure and Statistics}

Using the annotations obtained, we constructed two datasets: 1)~an IR dataset consisting of a collection of Kazakh Wikipedia passages, a list of questions, and pairs of question and passage IDs marked as either relevant or non-relevant (represented in TREC \emph{qrels} format), and 2)~a machine-reading dataset in SQuAD-like format that contains question-passage-answer triples. 

The statistics of the IR part can be found in Table~\ref{tab:stats}. 
All NQ questions were included in the \emph{training} set.
From the UNT annotation, we retained 548 questions for the \emph{development} set to ensure a roughly equal distribution of questions from each of the five subjects.
The remaining 1,929 questions formed the \emph{test} set.
As one can see, the total number of annotated queries is slightly below 6,000, while the number of passage-level annotations (both relevant and non-relevant) is almost 12,000. These figures are on par with the average numbers per language in the MIRACL dataset~\cite{zhang2022making}, although the share of relevant labels is higher in KazQAD. 

\begin{table}[!h]
\begin{center}
\begin{tabular}{lrrr}
\toprule
\textbf{Source} & \textbf{Q} & \textbf{P\textsuperscript{+}} & \textbf{P\textsuperscript{--}} \\
\midrule
NQ (train) & 3,487 & 3,893 & 3,558 \\
UNT (dev) & 548 & 769 & 229 \\
UNT (test) & 1,929 & 2,718 & 653 \\
\midrule
\textbf{Total} & \textbf{5,964} & \textbf{7,380} & \textbf{4,440} \\
\bottomrule
\end{tabular}
\caption{KazQAD statistics: Q -- annotated questions,  P\textsuperscript{+} and P\textsuperscript{--} -- relevant and non-relevant passages.}
\label{tab:stats}
\end{center}
\end{table}

\begin{table*}[t]
\centering
\begin{tabular}{lrrrr}
\toprule
& \multicolumn{3}{c}{\textbf{KazQAD}} & \multirow{2}{*}{\textbf{NQ\textsubscript{transl.}}} \\
\cmidrule{2-4}
& \textbf{train} & \textbf{dev} & \textbf{test} &  \\
\midrule
\textbf{\# question-passage pairs} & 3,163 & 764 & 2,713 & 61,606 \\
\textbf{\# unique questions} & 2,920 & 545 & 1,927 & 61,198 \\
\textbf{\# unique passages} & 1,993 & 697 & 2,137 & 53,204 \\
\textbf{\# short answers} & 3,826 & 910 & 3,315 & 71,242 \\
\textbf{avg. words per passage} & 72.9 & 105.1 & 107.1 & 72.7 \\
\textbf{avg. words per question} & 6.1 & 6.8 & 6.8 & 6.6 \\
\textbf{avg. words per answer} & 3.5 & 2.0 & 1.9 & 3.6 \\
\textbf{avg. question-passage LCS} & 11.6 & 12.9 & 13.0 & 13.1 \\
\bottomrule
\multicolumn{5}{l}{\textit{Note.} LCS stands for the longest common substring.}
\end{tabular}
\vspace{-0.25cm}
\caption{KazQAD and NQ translated statistics.}
\label{tab:data_stats}
\end{table*}

The statistics of the RC part of KazQAD are presented in Table~\ref{tab:data_stats}. 
Note that the number of unique questions in the RC part is slightly lower than in the IR part: for some passages that were considered relevant, the annotators could not extract short answers.
At the same time, each question can have more than one answer-bearing passage, which is reflected in the higher number of unique question-passage pairs. 
A relevant passage can contain several correct answers that cannot be extracted as a single span (e.g., in the case of list questions):

\begin{dialogue}
\speak{Q} \textit{What five countries border the Caspian Sea?}
\speak{A} \textit{Kazakhstan, Russia, Azerbaijan, Turkmenistan, Islamic Republic of Iran} (Translated)
\end{dialogue}

The UNT and NQ parts are rather similar in terms of \textit{question} length, but UNT \textit{answers} are significantly shorter. 
Note that the original UNT data comprise question-answer pairs, and most of the answers were kept unchanged by the annotators. 
The last row in the Table~\ref{tab:data_stats} estimates the similarity of questions and answer-bearing passages using the length of the longest common substring.\footnote{We employed the \texttt{difflib} library, see \url{https://docs.python.org/3/library/difflib.html}.} 
A high lexical overlap between the question and the corresponding context makes the task of locating the answer easier. 
High question-paragraph similarity was identified as a shortcoming of the SQuAD dataset, where the annotators generated questions based on presented Wikipedia paragraphs. 
Although a direct comparison between the languages is not precise due to morphological differences, we can surmise that the question-passage similarity in KazQAD is much lower than, for example, in English SQuAD and Russian SberQuAD/XQuAD~\cite{sberquad}.

\begin{table*}[ht]
\centering
\small
\begin{tabular}{p{.5\textwidth}|p{.45\textwidth}}
\toprule
\textbf{Kazakh} & \textbf{English translation} \\
\midrule 
\textbf{Q: } Әйелдер хоккейі қашан олимпиадалық спорт түріне айналды? & \textbf{Q:} When did women's hockey become an Olympic sport? \vspace{2pt} \\
\underline{1998 жылдан бастап} Олимпиадаларда әйелдер арасындағы хоккей турнирі де өткізіле бастады. & 
\underline{Since 1998}, a women's hockey tournament has also been held at the Olympics. \\

\midrule

\textbf{Q: }Қазақ халқынан шыққан алғашқы ғарышкер кім? &
\textbf{Q: } Who is the first Kazakh astronaut? \vspace{2pt}\\
\underline{Тоқтар Оңғарбайұлы Әубәкіров} (27 шілде 1946 жыл, Қарқаралы ауданы, Қарағанды облысы, Қазақстан) — қазақтан шыққан тұңғыш ғарышкер, ұшқыш, Кеңес Одағының Батыры (1988), Қазақстан Республикасының Халық қаһарманы (1995), техника ғылымының докторы (1998), профессор (1997), Қорқыт Ата атындағы Қызылорда мемлекеттік университетінің құрметті профессоры. Академик Е.А.Бөкетов атындағы Қарағанды университетінің Құрметті профессоры (3 мамыр 2022 жыл). &

\underline{Toktar Ongarbayuly Aubakirov} (July 27, 1946, Karkaraly district, Karaganda region, Kazakhstan) — the first Kazakh cosmonaut, pilot, Hero of the Soviet Union (1988), People's hero of the Republic of Kazakhstan (1995), doctor of technical sciences (1998), professor (1997), honorary professor of the Korkyt Ata Kyzylorda State University. Honorary professor of Karaganda University named after Academician Y. A. Boketov (May 3, 2022). \\  
\bottomrule
\end{tabular}
\caption{Examples of question-passage-answer triples from the MRC part of KazQAD (\underline{Underlined} are the annotated answers). The top example comes from the training set based on NQ: the question is machine-translated, while the passage comes from the Kazakh Wikipedia. The bottom example is an original Kazakh question (possibly rephrased by an annotator) from UNT, aligned with a passage from the Kazakh Wikipedia.}
\label{tab:examples}
\end{table*}

During our review of the obtained annotations, it came to our attention that there exist a notable number of lengthy answers, with 142 answers in the NQ part exceeding 100 characters. Upon closer analysis, it became apparent that the long answers can be attributed, in part, to the nature of the questions themselves:
\begin{dialogue}
\speak{Q} \textit{What is the function of albumin in the blood?}
\speak{A} \textit{forms a complex connection with vitamins, microelements, hormones and transports them throughout the body} (Translated)
\end{dialogue}
\noindent and, in part, to the inclusion of excessively detailed descriptions related to the answer entities:
\begin{dialogue}
\speak{Q} \textit{What part of the world is Greece located in?}
\speak{A} \textit{In the south-east of Europe, in the south of the Balkan Peninsula and on small islands in the Ionian, Mediterranean and Aegean seas adjacent to it.} (Translated)
\end{dialogue}

Table~\ref{tab:examples} shows two question-passage-answer triples from KazQAD: the first features the translated question from the English NQ, the second an original Kazakh question from UNT (albeit with slight manual changes from a multiple-choice question). 
These examples illustrate that UNT contains more `localised' questions (at least in the subsections corresponding to \textit{history of Kazakhstan} and \textit{Kazakh literature}). 
In this regard, we follow the recommendations formulated by the creators of multilingual QA datasets that the data should reflect the cultural, historical, and geographical context whenever possible~\cite{mkqa,tydiqa}.

\section{Methods and Baselines} \label{sec:baselines}
To showcase the difficulty of the dataset, we implemented and evaluated models for both RC and IR models.
Subsequently, we integrated retrieval and reading comprehension modules to evaluate the performance of a complete ODQA system.
Prior to discussing the experimental results, we provide an overview of the Transformer models employed as backbones for the IR, RC, or both systems.

\paragraph{Backbone Transformer Models} 

We use several Transformer-based models for our baseline solutions. 
Kazakh is featured in popular pre-trained multilingual models, such as mBERT~\cite{bert} and \mbox{XLM-R~\cite{xlm-r}}.
Multilingual models enable cross-lingual transfer learning, wherein the model is fine-tuned with data in one language and subsequently applied to another language.  
mBERT is a multilingual variant of BERT pre-trained on a mixture of the 104 largest Wikipedias.
mBERT has a shared 110,000 vocabulary, 12 hidden layers with 12 attention heads.
XLM-R~\cite{xlm-r} is a multilingual variant of RoBERTa, which in turn follows the learning regime of BERT with some optimizations, trained on a 100-language corpus from Common Crawl.
We use the \textit{Base} version of the model with a 250,000-token vocabulary and 12 hidden layers with 12 attention heads.
This model outperforms mBERT on a variety of tasks, especially in the case of low-resourced languages.
XLM-V~\cite{xlm-v} is a successor to XLM-R with an extended vocabulary aimed at overcoming the \textit{vocabulary bottleneck}.
The architecture of XLM-V is similar to that of XLM-R, but has a vocabulary of one million tokens, allowing for more meaningful tokenisation.
XLM-V outperforms its predecessor on multilingual natural language inference (NLI), NER, and QA tasks.
Kaz-RoBERTa\footnote{\url{https://huggingface.co/kz-transformers/kaz-roberta-conversational}} is a monolingual model trained on a collection of Kazakh texts from Common Crawl, the Leipzig Corpora Collection, the OSCAR corpus, as well as Kazakh books and news, with a total of about two billion tokens.
The model has a vocabulary of 52,000 tokens and features six hidden layers, in contrast to the 12 layers found in other Transformer-based models in the study, with 12 attention heads.

It is important to note that   
the quantity of language data available for Kazakh is relatively small compared to the entire training corpora of the aforementioned multilingual models.
For example, the number of Wikipedia Kazakh articles used for mBERT pre-training, is about 30 times smaller than the number of English articles.
Moreover, because Kazakh articles are typically shorter than English ones, the gap is even larger when the number of tokens is taken into account. 
XLM-R was pre-trained on a larger multilingual corpus derived from Common Crawl,
in which Kazakh is 44th out of 100 languages by the volume of the training data. 
In terms of the number of tokens, the Kazakh subcorpus is 117 times smaller than the English subcorpus and 5.8 times smaller than the Turkish subcorpus~\cite{xlm-r}. 
The size of the language data used for pre-training and the size of the model vocabulary allocated to the language are among the main factors that impact the quality of the model fine-tuned on downstream tasks and cross-lingual learning performance~\cite{lauscher2020zero,xquad}.

\paragraph{IR baselines} 
We implemented a classic filter-and-refine multi-stage retrieval
\emph{pipeline} \cite{DBLP:conf/sigir/MatveevaBBLW06}, 
where top-$k$ (k=$10^4$) candidate documents obtained with a fast BM25 ~\cite{robertson2009probabilistic}  ranker 
are re-ranked using a more accurate (but slower) ranker. Our experiments are conducted using the FlexNeuART~\cite{boytsov2020flexible} framework.

Our candidate generator relies on whitespace tokenisation, which includes lowercase conversion and punctuation removal.
The index is built over the concatenation of the title of the Wikipedia page and the passage text. 
Using this candidate generator, we implemented two non-neural rankers that employ a simple linear learning-to-rank (LETOR) algorithm, namely coordinate ascent, to combine several scores.
Linear weights are found using a subset of the training set.
The first non-neural ranker is a multi-field BM25 baseline, which includes separate scores for the title
and the passage text. 
Each of these fields undergoes dual tokenisation: whitespace tokenisation and mBERT tokenisation, resulting in a total of four fields.

The second ranker includes scores for the monolingual 
IBM Model 1~\cite{brown-etal-1993-mathematics}, a lexical translation model that requires a parallel corpus. 
As a parallel corpus, we use questions paired with answer-bearing text snippets.
To find these snippets, we use BM25 to retrieve paragraphs and then select substrings that contain the answer text. 
We restrict the snippets to include a maximum of five additional words on both the left and right sides of the answer.

The top-100 results produced by the best non-neural ranker (from $10^4$ BM25 candidates) are re-ranked with a standard cross-encoding BERT ranker \cite{DBLP:journals/corr/abs-1901-04085},
which uses XLM-R or mBERT as its backbone model. 
The ranker is first pre-trained on English MS MARCO (passage corpus) \cite{DBLP:journals/corr/abs-2003-07820} and then fine-tuned on the KazQAD data.
Although KazQAD has manually annotated relevance passages, we found that better results are obtained when training with weakly supervised data \cite{karpukhin2020dense}.
Similar to the creation of the parallel corpus, we first retrieved paragraphs using BM25. 
The paragraphs containing the answer text were deemed to be relevant. 
Note, however, that, the evaluation results are based on human-provided relevance data.

We evaluated both zero-shot and fine-tuned rankers.
According to the results in Table~\ref{tab:ir}, both XLM-R and mBERT zero-shot rankers have strong performance, which improves only by about 5\% after fine-tuning using the KazQAD data.
Thus, our best pipeline uses the fine-tuned XLM-R ranker.

\begin{table}[h]
\fontsize{9.8}{10.8}\selectfont
\setlength\tabcolsep{0.003cm}
\begin{tabularx}\columnwidth{lccc}
\toprule
\textbf{IR pipeline}  &    \textbf{ NDCG@10}  &  \textbf{MRR}  &  \textbf{R@100} \\\midrule
\multicolumn{4}{c}{candidate generators} \\\midrule
BM25 (title+text)   &  0.1446 & 0.1443 & 0.4491 \\\midrule
\multicolumn{4}{c}{non-neural/classic re-rankers} \\\midrule
BM25 (multi-field)         &  0.1992 & 0.1961 & 0.5559 \\
BM25+Model1 (multi-field)  &  0.2735 & 0.2723 & \textbf{0.6478} \\\midrule
\multicolumn{4}{c}{best classic + neural re-rankers} \\\midrule
mBERT (zero-shot)          &  0.345 & 0.3344 & \textbf{0.6478} \\
mBERT (fine-tuned)         &  0.3672 & 0.3628 & \textbf{0.6478} \\
XLM-R (zero-shot)          &  0.3764 & 0.3654 & \textbf{0.6478} \\
XLM-R (fine-tuned)         &  \textbf{0.3892} & \textbf{0.3822} & \textbf{0.6478} \\
\bottomrule
\end{tabularx}
\caption{Effectiveness of baseline IR systems.}
\vspace{-0.3cm}
\label{tab:ir}
\end{table}

\paragraph{RC baselines} We conducted reading comprehension experiments using three pre-trained Transformer-based models: Kaz-RoBERTa, XLM-R, and XLM-V.
We fine-tuned these models in several scenarios: on 61,000 machine-translated question-paragraph pairs (NQ\textsubscript{transl.}), on 69,000 English question-passage pairs (NQ\textsubscript{en}) that were not used in the creation of KazQAD, on the KazQAD train set, and on their combinations.

To fine-tune the models, we used scripts and parameters from the official Hugging Face repository.\footnote{\url{https://github.com/huggingface/transformers/tree/main/examples/pytorch/question-answering}}
For each training data configuration, we tried different numbers of epochs and selected the best model based on the KazQAD validation set. In scenarios with two data sources, we first selected the best model that was tuned on the first dataset and then tuned it on the second dataset.

\begin{table}[h]
\setlength\tabcolsep{0.15cm}
\begin{tabularx}\columnwidth{llcc}
\toprule
\textbf{Model} & \textbf{Train} & \textbf{EM} & \textbf{F1}\\
\midrule
Kaz-RoBERTa    & NQ\textsubscript{transl.}        & 20.86 & 31.82 \\
     & KazQAD                           & 10.25 & 18.31 \\
     & NQ\textsubscript{transl.}+KazQAD & 22.45 & 35.46 \\
\midrule
XLM-RoBERTa    & NQ\textsubscript{en}             & 32.62 & 47.18 \\
     & NQ\textsubscript{transl.}        & 32.33 & 47.75 \\
     & KazQAD                           & 24.33 & 37.49 \\
     & NQ\textsubscript{en}+KazQAD      & 37.38 & 52.67 \\
     & NQ\textsubscript{transl.}+KazQAD & 36.56 & 51.87 \\
\midrule
XLM-V          & NQ\textsubscript{en}+KazQAD      & \textbf{38.52} & \textbf{54.18} \\
     & NQ\textsubscript{transl.}+KazQAD & 38.48 & 52.78 \\
\bottomrule
\end{tabularx}
\caption{Effectiveness of baseline MRC models.}
\label{tab:mrc}
\end{table}

The results for different models and training set configurations are presented in Table~\ref{tab:mrc}. 
We found that the monolingual Kaz-RoBERTa model performs dramatically worse than the multilingual models. 
This may be due to the fact Kaz-RoBERTa is a smaller model with only six hidden layers.

Another surprising result was that the multilingual models fine-tuned on English and on the translated NQ show similar performance. 
Furthermore, when we trained XLM-R on the manually annotated KazQAD training set, we obtained much lower scores.
However, additional fine-tuning on KazQAD improved both EM and F1 by about five points.
Using XLM-V instead of XLM-R further increased EM and F1 by about two points.

\begin{table}[h]
\setlength\tabcolsep{0.18cm}
\centering
\begin{tabularx}\columnwidth{lcrr}
\toprule
\textbf{IR-pipeline} & \textbf{Fusion} & \textbf{EM} & \textbf{F1} \\
\midrule
\multicolumn{4}{c}{non-neural/classic retriever}\\\midrule
BM25 (title+text)          & \xmark &  7.7 & 12.6 \\
BM25 (multi-field)         & \xmark &  8.5 & 14.3 \\
BM25+Model1 (multi-field)  & \xmark &  12.4 & 19.8 \\\midrule
\multicolumn{4}{c}{BM25 (title+text) + neural ranker} \\\midrule
XLM-R (fine-tuned)   & \xmark & 15.2 & 25.2 \\\midrule
\multicolumn{4}{c}{best classic + neural ranker} \\\midrule
XLM-R (fine-tuned)   & \xmark &  17.3 & 28.1 \\
XLM-R (fine-tuned)   & \cmark &  \textbf{17.8} & \textbf{28.7} \\

\bottomrule
\end{tabularx}
\caption{ODQA baselines (XLM-V reader).}
\label{tab:odqa}
\end{table}

\paragraph{ODQA baselines} 
We test our best reader (XLM-V model) in the ODQA setting~\cite{hirschman2001natural},
where the reader extracts answers from top-$k$  passages produced by an IR system (we use $k=100$). 
Because each passage generates an answer, there should be a final answer aggregation and/or selection process.
We tried two simple approaches: using the top-1 retrieved document and \emph{fusion} of the retriever and reader scores (with
a subsequent selection of the top-fusion-score answer)~\cite{yang-etal-2019-end-end}.
Fusion scores were estimated using the development set.

In Table~\ref{tab:odqa}, we show QA accuracy for progressively improving retrieval systems. 
Using BM25 alone for concatenated title and text fields produced a very low F1 score of 12.6, which increased 1.6 times when we used a strong classic IR pipeline BM25+Model1 (multi-field).
A further 1.4 times increase was achieved when we additionally applied a neural ranker.

The neural ranker was quite helpful, even if we used it on top of our weakest IR system (see scores for ``BM25 (title + text) + neural ranker'').
However, the resulting score was still 1.1 times lower compared to our top-system without fusion.
This underscores the need for a carefully optimised retriever.
Unfortunately, fusing a reader and retriever scores was only marginally helpful.

\paragraph{ChatGPT} We fed questions from the KazQAD test set to the OpenAI model \texttt{gpt-3.5-turbo-0125}\footnote{\url{https://platform.openai.com/docs/models/gpt-3-5-turbo}} preceded by a simple prompt in English: ``\textit{The question is in the Kazakh language. Provide a short and concise answer in Kazakh.}''
Given the potential verbosity of ChatGPT responses, we employed the \textit{recall} of lemmatised words as an evaluation metric. In instances where multiple answers were available in KazQAD, we concatenated them for consistency.
Tokenisation and lemmatisation were carried out using the Stanza library~\cite{qi-etal-2020-stanza}.
An additional evaluation criterion was the length of the longest common substring between the responses generated by ChatGPT and those provided in KazQAD.
Of the 1,927 test questions, 173 (9\%) elicited responses with a recall score of 0.5 or higher, with only 86 (4.5\%) achieving a perfect recall score.
Manual examination of the results revealed instances where ChatGPT provided conspicuously inaccurate answers, such as incorrectly identifying ``Shakespeare'' as Ablai Khan's successor instead of the correct answer, ``Uali''.
Furthermore, inaccuracies were observed in responses such as the erroneous identification of Kazakh as the language of inter-ethnic communication in Kazakhstan instead of Russian. However, despite these errors, the model occasionally gave correct answers, albeit with small variations.
For example, while the KazQAD answer for \textit{Кеңес елінің басшысы М. С. Горбачевтің жүргізген реформасының аты қандай?} (What is the name of M. S. Gorbachev's reform?) is \textit{Қайта құру кезеңі}, the original Russian term \textit{Перестройка}, returned by ChatGPT, is also considered correct, which illustrates variations in answer presentation.
Differences in character representation were also observed, such as the use of Cyrillic characters instead of Latin characters (e.g., [\textit{витамин} (vitamin)] D in KazQAD vs. \textit{Д} by ChatGPT), or the use of the Cyrillic dotted \textit{і} where the Cyrillic \textit{и} is expected (e.g., \textit{Галилей}, \textit{Италия}, \textit{испандық} in KazQAD vs. \textit{Галілей}, \textit{Італия}, \textit{іспандық} by ChatGPT ), which overall reflects the the markedly inferior performance of OpenAI's model in answering factual questions in Kazakh and underscores the importance of manually annotated datasets over machine-translated data.

\section{Conclusion}

We introduce \textbf{KazQAD}: a Kazakh IR, RC, and ODQA dataset, which is one of the few annotated NLP/IR resources for the Kazakh language. 
The annotated data are publicly available together with a collection of Kazakh Wikipedia passages and a machine-translated subset of the Natural Questions dataset.
In preparing the dataset, we tried to reuse existing resources and reduce the cost of manual annotation. 
We also created a set of baselines for retrieval, reading comprehension, and ODQA tasks. 
We hope that both our dataset preparation approach and the results presented herein will contribute to research and applications concerning Kazakh and other low-resource languages.

\section{Ethics Statement}
We ensured that the annotators involved in this study received fair compensation for their work, 
with the workload staying well within the bounds of a standard working day.

\section{Bibliographical References}\label{reference}

\bibliographystyle{bibliography_style}
\bibliography{bibliography}

\end{document}